%% file: main.tex
\documentclass[runningheads]{llncs}
\usepackage[utf8]{inputenc} 
\usepackage[T1]{fontenc}    
\usepackage{hyperref}       
\usepackage{url}            
\usepackage{booktabs}       
\usepackage{amsfonts}       
\usepackage{nicefrac}       
\usepackage{microtype}      

\usepackage{mathops}
\usepackage{cancel}
\usepackage{graphicx}
\usepackage{subfigure}
\usepackage{marvosym}
\usepackage{wrapfig}
\usepackage{fixmath}
\usepackage{ulem}
\usepackage[draft=true]{minted}
\usepackage{cprotect}

\newcommand{\g}{$\sigma$ }
\newcommand{\mug}{$\sigma_0$ }
\newcommand{\sigmag}{$\sigma_1$ }
\begin{document}
\title{Universal Representation for Code}
%
%
\author{Linfeng Liu\inst{1}\thanks{To appear on PAKDD 2021. Work done while the author was an intern at AWS.}$^{(\textrm{\Letter})}$ \and
Hoan Nguyen \inst{2} \and
George Karypis \inst{2} \and
Srinivasan Sengamedu \inst{2}}
%
\authorrunning{L. Liu et al.}
%
\institute{Tufts University, Medford MA 02155, USA \\
\email{linfeng.liu@tufts.edu} \and
Amazon Web Services, Seattle WA 98109, USA\\
\email{\{hoanamzn, gkarypis, sengamed\}@amazon.com}}

\maketitle              
\begin{abstract}
\input{abstract.tex}
\end{abstract}

\section{Introduction}
\input{introduction.tex}

\section{Preliminary}
\input{preliminary.tex}

\section{Code Graph}
\input{graph.tex}

\section{Method}
\input{method.tex}

\section{Experiment}
\input{experiment.tex}

\section{Conclusion}
\input{conclusion.tex}

\bibliographystyle{ieeetr}
\bibliography{reference}

\end{document}

%% file: abstract.tex
Learning from source code usually requires a large amount of labeled data. Despite the possible scarcity of labeled data, the trained model is highly task-specific and lacks transferability to different tasks. In this work, we present effective pre-training strategies on top of a novel graph-based code representation, to produce \textit{universal} representations for code.
Specifically, our graph-based representation captures important semantics between code elements (e.g., control flow and data flow). We pre-train graph neural networks on the representation to extract universal code properties. The pre-trained model then enables the possibility of fine-tuning to support various downstream applications. We evaluate our model on two real-world datasets -- spanning over 30M Java methods and 770K Python methods. Through visualization, we reveal discriminative properties in our universal code representation. By comparing multiple benchmarks, we demonstrate that the proposed framework achieves state-of-the-art results on method name prediction and code graph link prediction.

\keywords{Code representation \and Graph neural network  \and Pre-training.}

%% file: introduction.tex
Analysis of software using machine learning approaches has several important applications such as identifying code defects~\cite{dinella2019}, improving code search~\cite{cambronero2019}, and improving developer productivity~\cite{raychev2014}. One common aspect of any code-related application is that they learn code representations by following a two-step process. The first step takes code snippets and produces a \textit{symbolic} code representation using program analysis techniques. The second step uses the symbolic code representation to generate \textit{neural} code representations using deep learning techniques.

Symbolic representations need to capture both syntactic and semantic structures in code. Approaches to generating symbolic representations can be categorized as sequence-, tree-, and graph-based. Sequence-based approaches represent code as a sequence of tokens and only capture the shallow and textual structures of the code~\cite{allamanis2017}. Tree-based approaches represent the code via abstract syntax trees (ASTs) \cite{tbcnn} that highlight structural and content-related details in code. However, some critical relations (e.g., control flow and data flow), which often impact machine learning models' success in abstracting code information, are not available in trees. Graph-based approaches augment ASTs with extra edges to partially represent the control flow and the data flow ~\cite{dinella2019,allamanis2017,hu2020}.
Depending on the type of symbolic representation used, the approaches for generating the neural code representations are either sequence-based~\cite{raychev2014,hindle2012} or graph-based~\cite{allamanis2017,hu2020} neural network models.
However, these works are generally task-specific, making it hard to transfer the learned representations to other tasks. In addition, the scarcity of labeled data may cause insufficient training in deep learning models.

\begin{figure*}[t]
    \centering
    \includegraphics[width=0.93\textwidth]{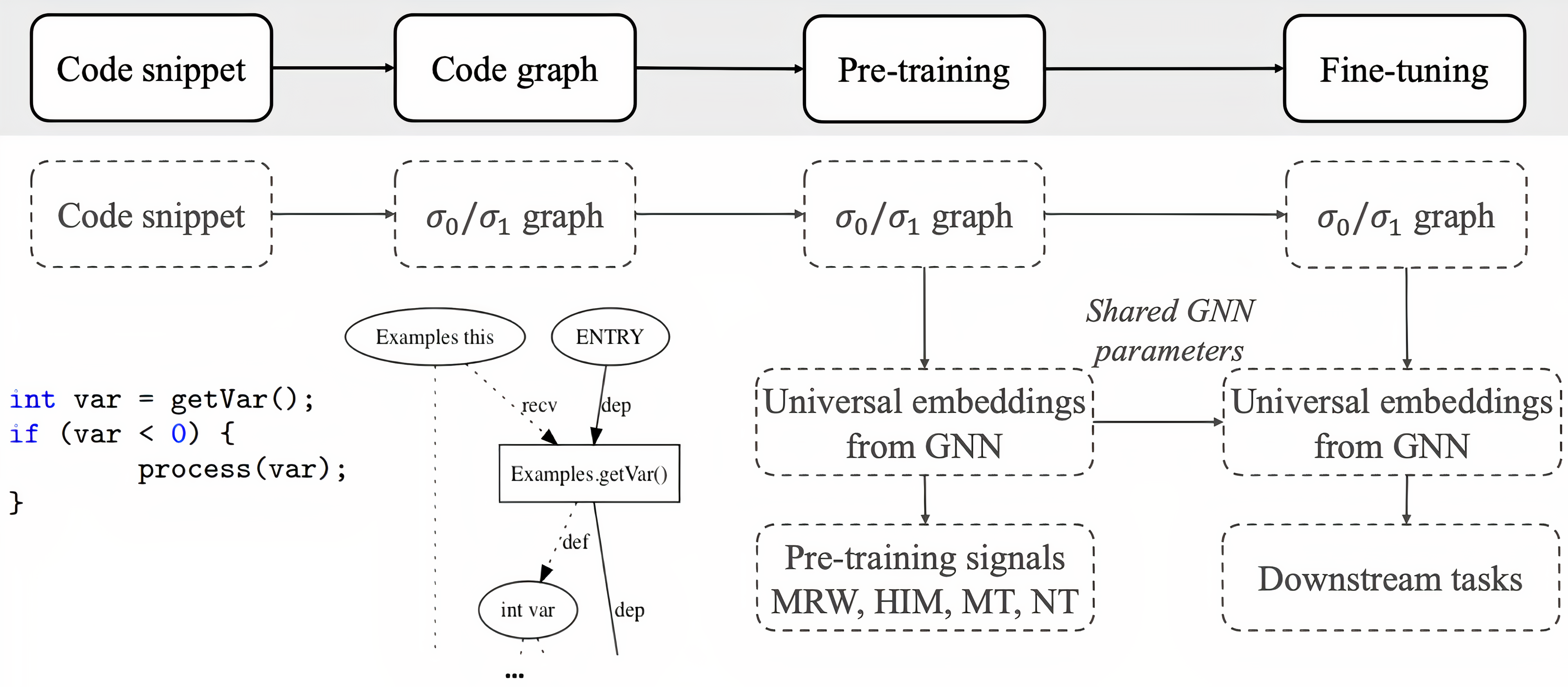}
    \caption{Model pipeline. The $\sigma_0/\sigma_1$ graph is defined in Section 3, and the pre-training signals are defined in Section 4.}
    \label{model-overview}
\end{figure*}
In this work, we touch upon all the three aspects of ML-based code analysis: symbolic code representation, task-independent neural code representation, and task-specific learning. Fig.~\ref{model-overview} gives an overview of our approach. For symbolic code representation, we explore two alternatives and show that symbolic code representation (called $\sigma_0/\sigma_1$ graph) which captures richer relations leads to better performance in downstream tasks. For neural code representation, we specialize a recently proposed universal representation for graphs, PanRep~\cite{ioannidis2020}, to code graphs. And, finally, we explore two tasks to demonstrate the effectiveness of the learned representations: method name prediction (for Python and Java) and link prediction (for Java).
Our proposed method consistently improves the prediction accuracy across all experiments. 

To summarize, the contributions of this work are as follows:
\begin{itemize}
    \item We introduce a fine-grained symbolic graph representation for source code, and adapt to 29M Java methods collected from GitHub.
    \item We present a pre-training framework that leverages the graph-based code representations to produce universal code representations, supporting various downstream tasks via fine-tuning.
    \item We combine the graph-based representation and the pre-training strategies to go beyond code pre-training with sequence- and tree-based representations.
\end{itemize}

%% file: preliminary.tex
\parhead{Notation} Let $G=\{\calV, \calE\}$ denote a \textit{heterogeneous} graph with $|\calT|$ node types and $|\calR|$ edge types. $\calV = \{\{\calV^t\}_{t \in \calT}\}$ represents the node set, and $\calE = \{\{\calE^r\}_{r \in \calR}\}$ represents the edge set. Each node $v_i^t \in \calV^t$ is associated with a feature vector. Throughout the paper, we often use ``representation'' and ``embedding'' interchangeably unless there is any ambiguity.

\subsection{Graph Neural Networks}
Graph neural networks (GNNs) learn representations of graphs \cite{wu2020}. A GNN typically consists of a sequence of $L$ graph convolutional layers. Each layer updates nodes' representation from their direct neighbors. By stacking multiple layers, each node receives messages from higher-order neighbors. In this work, we utilize the relational graph convolutional network (RGCN)~\cite{schlichtkrull2018} to model our heterogeneous code graphs. RGCN's update rule is given by
\begin{align}
    \bh_{i}^{(l+1)} = \phi\left(\sum_{r \in \calR} \sum_{n \in \calN_i^r} \frac{1}{c_{i,r}}\bh_n^{(l)} \bW_r^{(l)} \right), \nonumber
\end{align}
where $\calN_i^r$ is the neighbor set of node $i$ under edge type $r$, $c_{i,r}$ is a normalizer (we use $c_{i,r}=|\calN_i^r|$ as suggested in \cite{schlichtkrull2018}), $\bh_i^{(l)}$ is the hidden representation of node $i$ at layer $l$, $\bW_r^{(l)}$ are learnable parameters, and $\phi(\cdot)$ is any nonlinear activation function. Usually, $\bh_i^{(0)}$ is initialized as node features, and $\bh_{i}^{(L)}$ (the representation at the last layer) is used as the final representations.

\subsection{Pre-training for GNNs and for Source Code}
Recently, there is a rising interest in pre-training GNNs to model graph data~\cite{hu2019strategies,jin2020}.
To pre-train GNNs, most works encourage GNNs to capture graph structure information (e.g., graph motif) and graph node information (e.g., node feature). PanRep~\cite{ioannidis2020} further extends GNN pre-training to heterogeneous graphs.

Pre-training on source code has been studied in \cite{kanade2020,feng2020,svyatkovskiy2020}. However, these models build upon sequence-based code representations and fail to encode code's structural information explicitly. We differ from these works, by pre-training on a novel graph-based code representation to capture code's structural information.

%% file: graph.tex
Previous Machine Learning (ML) models \cite{allamanis2017,hu2020,alon2018} are largely based on ASTs to reflect structural code information. Though ASTs are simple to create and use, they have tree-based structures and do not capture control flow and data flow relations. Here, the control flow represents the order of the execution and the data flow represents the flow of data along the computation. For example, to represent a loop snippet, ASTs cannot naturally use an edge pointing from the end of the program statement to the beginning of the loop. In addition, the relation between the definition and uses of a variable is not captured in ASTs. In this work, we represent code as graphs to efficiently capture both control flow and data flow between program elements. We call our code graphs as \mug graphs and \sigmag graphs. The \mug graphs are related to classical Program Dependence Graphs (PDGs)~\cite{pdg:1987}. The \sigmag graphs build upon the \mug graphs and include additional syntactic and semantic information (detailed in section \ref{sec-sigma}). Our experiments show that ML models using the \sigmag graphs achieve better prediction accuracy than using the \mug graphs.
\begin{wrapfigure}{r}{0.38\textwidth}
  \begin{center}
    \includegraphics[scale=0.5]{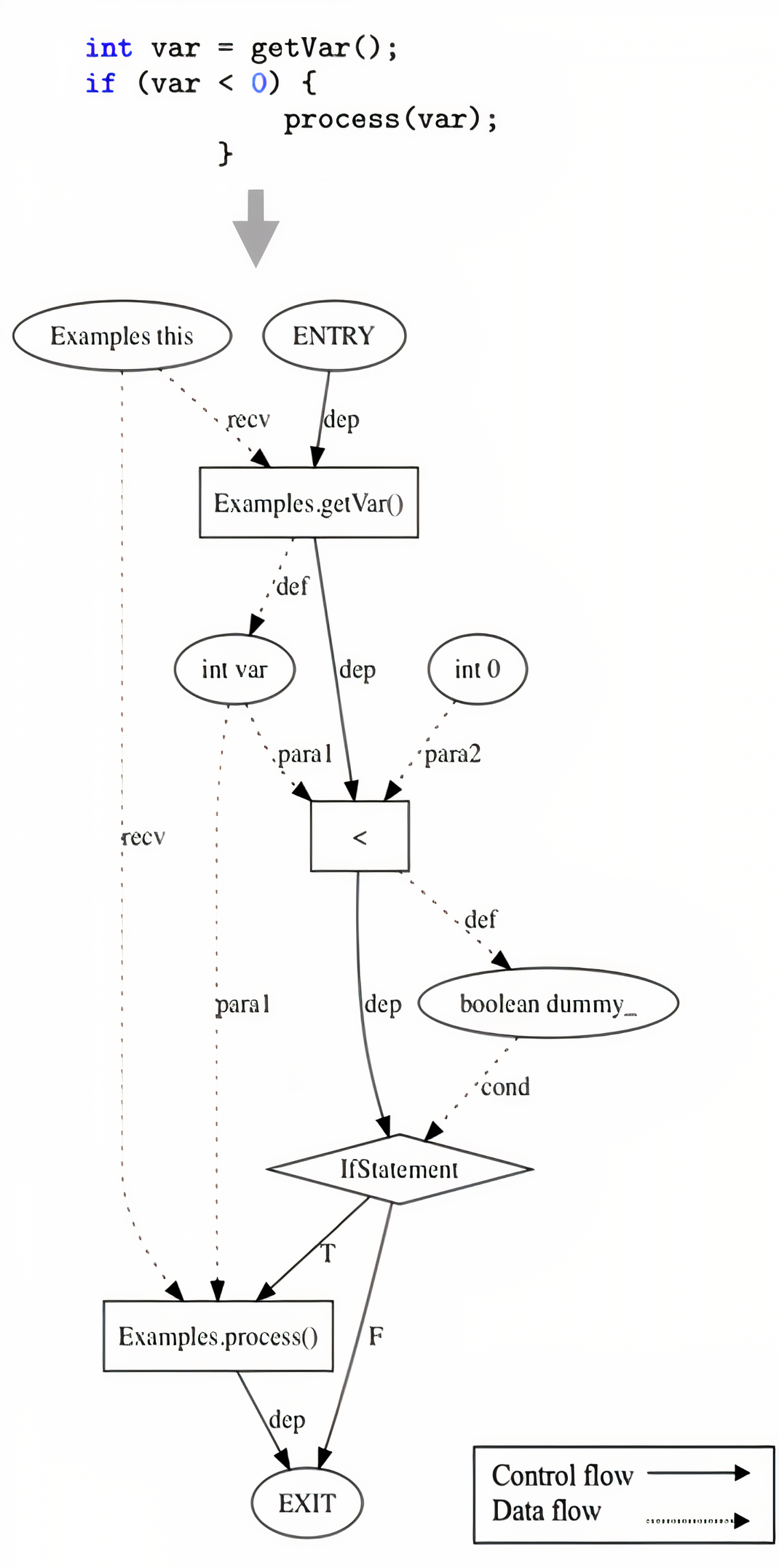}
  \end{center}
  \vspace{-5mm}
  \caption{An example of \mug graph.}
    \label{fig:mu-graph}
\vspace{-15mm}
\end{wrapfigure}

\subsection{The \mug Graph}

The \mug graphs, which relate to PDGs, are used for tasks such as detection of security vulnerabilities and identification of concurrency issues.
In \mug graphs, nodes represent different kinds of program elements including data and operations; edges represent different kinds of control flow and data flow between program elements. We showcase a \mug graph in Fig. \ref{fig:mu-graph}.

Both nodes and edges in the \mug graph are typed. Specifically, we have five node types: entry, exit, data, action, and control nodes. Entry and exit nodes indicate the control flow entering and exiting the graph. Data nodes represent the data in programs such as variables, constants and literals. Action nodes represent the operations on the data such as method calls, constructor calls and arithmetic/logical operations, etc. Control nodes represent control points in the program such as branching, looping, or some special code blocks such as catch clauses and finally blocks. We have two edge types: control and data edges. Control edges represent the order of execution through the programs and data edges represent how data is created and used in the programs.
Examples of edge types include parameter edges which indicate the data flow into operations and throw edges which represent the control flows when exceptions are thrown.

\subsection{The \sigmag Graph} \label{sec-sigma}
One limitation of \mug graph is that the downstream modules perform additional analysis such as control dependence and aliasing is not reflected in the \mug graph explicitly and requires machine learning models to infer it.
We propose \sigmag graph as an augmentation of the \mug graph, which is enhanced by additional information.
Specifically, \sigmag graph attaches AST node types to nodes in the graph. AST node types capture syntactic information (e.g., InfixOperator) provided by the parser. Higher-order semantic relations such as variable usage (e.g., FirstUse/LastUse), node aliasing, and control dependence are also included as graph edges.

\subsection{The Heterogeneous Code Graph} \label{graph-label}
The proposed \mug and \sigmag graphs are heterogeneous graphs, i.e. nodes and edges have types and features. Nodes are categorized into five types: \verb+entry+, \verb+exit+, \verb+data+, \verb+action+, and \verb+control+. Node features are attributed by their names. Edge types are the same as edge features, which are defined by their functionalities. Below we first describe node features, followed by edge features.

Entry nodes have feature \verb+ENTRY+ and exit nodes have feature \verb+EXIT+. Features of control nodes are their corresponding keywords such as \verb+if+, \verb+while+, and \verb+finally+. Features of variables are types, and the variable names are ignored. For example, \verb+int.x+ $\rightarrow$ \verb+int+; \verb+String.fileName+ $\rightarrow$ \verb+String+. Method names contain the method class, method name, and parameter class. For example, \verb+Request.setConnectionKeepAlive#boolean#+. Here, \verb+Request+ is the method class, \verb+setConnectionKeepAlive+ is the method name, and \verb+boolean+ is the parameter class.

Features of control edges and data edges are defined separately. There are two kinds of control edges: normal control edges and exception control edges. 
A normal control edge is a directed edge that connects two control or action nodes; defined as \verb+dep+. See Fig. \ref{fig:mu-graph}.
An exception control edge connects an action node to a control node to handle an exception that could be thrown by the action; defined as \verb+throw+. Data edges have five features: \verb+receiver+, \verb+parameter+, \verb+definition+, \verb+condition+, and \verb+qualifier+. Examples include \verb+receiver.call()+ as a receiver edge; \verb+call(param)+ as a parameter edge, and \verb+foo=bar()+ as a definition edge.

\subsection{Corpus-level Graphs} \label{corpus-level-graph}
In a typical ML application, the corpus consists of a collection of packages or repositories. Repositories contain multiple files or classes. Classes contain methods. The \mug and \sigmag graphs are at \textit{method-level}. Corpus-level graphs are a collection of method-level $\sigma_0$/\sigmag graphs. Let \g refer to either \mug or \sigmag for notational ease.

%% file: method.tex
We propose a new model, \underline{Uni}versal \underline{Co}de \underline{R}epresentation via GN\underline{N}s (UniCoRN), to produce universal code representations based on \g graphs. UniCoRN has two components. First, a GNN encoder that takes in \g graphs and generates node embeddings. Second, pre-training signals that train the GNN encoder in an unsupervised manner. By sharing the same GNN encoder across all \g graphs, the learned embeddings reveal universal code properties. Below, we show our design of pre-training signals to help UniCoRN efficiently distill universal code semantics. The instantiation of the GNN encoder is given in the experiment.

\subsection{Pre-training Signals}
\parhead{Metapath Random Walk (MRW) signal.}
A MRW is a random path that follows a sequence of edge types. We assume node pairs in the same MRW are proximal to each other; accordingly, they should have similar embeddings. For example, for a MRW with nodes [\verb+Collection.iterator()+, \verb+Iterator+,  \verb+Iterator.hasNext()+] and edges [\verb+definition+, \verb+receiver+], nodes \verb+Iterator+ and \verb+Iterator.hasNext()+ are expected to have similar embeddings. Formally, the signal is defined as
\begin{align}
    \calL_{MRW} = \sum_{v, v' \in \calV} \log \left( 1 + \exp \left( -y \times \bh_v^t \bW^{t, t'} \bh_{v'}^{t'} \right) \right),
\end{align}
where $\bh_v^t$ and $\bh_{v'}^{t'}$ are embeddings for nodes $v$ and $v'$ with node types $t$ and $t'$. $\bW^{t, t'}$ is a diagonal matrix weighing the similarity between different node types. $y$ equals 1 as positive pairs if $v$ and $v'$ are in the same MRW, otherwise $y$ equals -1 as negative pairs. During training, we sample 5 negative pairs per positive pair.

\vspace{5pt}
\parhead{Heterogeneous Information Maximization (HIM) signal.}
Nodes of the same type should reside in some shared embedding space, encoding their similarity. On the other hand, nodes of different types, such as control nodes and data nodes, ought to have discriminative embedding space as they are semantically different. However, standard GNNs fail to do so with only local message propagation. Following \cite{ioannidis2020}, we use a HIM signal to encode these properties:
\begin{align}
        \calL_{HIM} =  \sum_{t \in \calT} \sum_{v \in \calV^t} \Big( 
            \log \big( \phi(\bh_v^\top \bW \bs^t) \big) + 
            \log \big( 1 - \phi(\Tilde{\bh}_v^\top \bW \bs^t) \big) 
        \Big).
\end{align}
Here, $\bs^t= \frac{1}{|\calV^t|} \sum_{v \in \calV^t}  \bh_v$ is a global summary of nodes typed $t$, $\phi(\cdot)$ is a sigmoid function, and $\phi(\bh_v^\top \bW \bs^t)$ quantifies the closeness between a node embedding $\bh_v$ and a global summary $\bs^t$. Negative samples $\tilde{\bh}_v$ are obtained by first row-wise shuffling input node features then propagating through the GNN encoder~\cite{ioannidis2020}.

\vspace{5pt}
\parhead{Motif (MT) signal.}
Code graph has connectivity patterns. For example, node \verb+ENTRY+ has only one outgoing edge; node \verb+IfStatement+ has True and False branch.
\begin{wrapfigure}{r}{0.34\textwidth}
\vspace{-10mm}
  \begin{center}
    \includegraphics[scale=0.09]{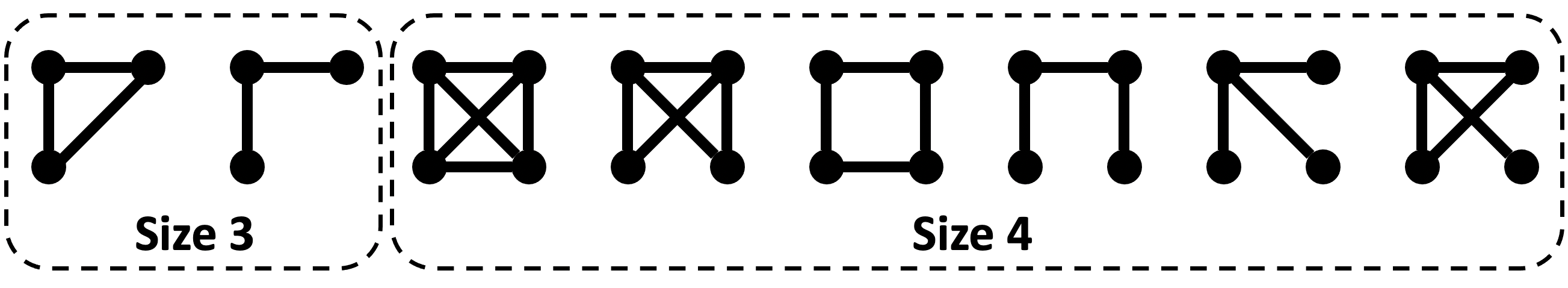}
  \end{center}
  \vspace*{-7mm}
  \caption{Motifs sized 3 and 4.}
  \vspace*{-7mm}
  \label{fig-motif34}
\end{wrapfigure}
Such structural patterns can be captured in graph motifs, see Fig.~\ref{fig-motif34}.
With this observation, we pre-train GNNs with MT signal to generate structure-aware node embeddings for code graphs. Formally, we aim to recover the ground truth motif around each node, $\bm{m}_v$, from the node embedding $\bh_v$ using an approximator $f_{MT}(\cdot)$ (we use a two-layer MLP),
\begin{align}
    \calL_{MT} = \sum_{v \in \calV} ||\bm{m}_v - f_{MT}(\bh_v)||_2^2.
\end{align}
The ground truth $\bm{m}_v$ is obtained using a fast motif extraction method \cite{ahmed2017}.

\vspace{5pt}
\parhead{Node Tying (NT) signal.}
The corpus-level graph (Cf. section \ref{corpus-level-graph}) contains many \textit{duplicate nodes} that have the same feature (e.g., two \verb+ENTRY+ nodes will be induced from two methods). 
These duplicate nodes serve as anchors to imply underlying relations among different graphs.
We divide duplicate nodes into two categories: strict equality and weak equality. Strict equality refers to duplicate nodes whose semantic meaning should be invariant to their context, including keywords (\verb+if+, \verb+while+, \verb+do+ ...), operators (\verb+=+, \verb+*+, \verb+<<+ ...), entry and exit nodes. Duplicate nodes of strict equality will have the same embedding in all \g graphs. We keep a global embedding matrix to maintain their embeddings. Weak equality refers to other duplicate nodes whose semantic meaning can be affected by their context. For example, two \verb+foo()+ nodes in two methods, or two tied nodes due to the simple qualified types of one or two nodes\footnote[1]{We use simple types instead of fully qualified types since we create graphs from source files and not builds. In this case, types are not fully resolvable.}. We use NT signal to encourage duplicate node of weak equality to have similar embeddings:
\begin{align}
    \calL_{NT} =\sum_{k \in \calK} \textrm{ave}(\{||\bh_v - \bg^k||_2^2\}, v \text{ has feature } k),
    \label{eq-nt}
\end{align}
where $\calK$ is the set of distinct node features (exclude strict equality nodes), $\textrm{ave}(\cdot)$ is an average function, and $\bg^k = \textrm{ave}(\{\bh_v\}, v \text{ has label } k)$ is a global summary of nodes featured $k$. In \eqref{eq-nt}, we first group nodes featured $k$, followed by computing the group centers $\bg^k$, then minimize the distance between nodes to their group centers.

\vspace{5pt}
\parhead{Pre-training objective.} 
We combine the four pre-training signals to yield a final objective:
\begin{align}
    \calL = \omega_{1}\calL_{MRW} + \omega_{2}\calL_{HIM} + \omega_{3}\calL_{MT} + \omega_{4}\calL_{NT},
    \label{eq-final-obj}
\end{align}
where $\omega_1,\ldots, \omega_4$ balance the importance of different signals. The objective resembles the objective in multi-task learning \cite{zhang2017survey}.

\subsection{Data Pre-processing and Fine-tuning}
\parhead{Numeric node features.}
The initial node features are strings (Cf. section \ref{graph-label}), which need to be cast into numeric forms before feeding into the GNN encoder. To this end, we first split each node's feature into subtokens based on the delimiter ``.''. Then, language models are used to get subtoken embeddings, in which we use FastText \cite{bojanowski2017}. Finally, we use average subtoken embeddings as the node's numeric feature.

\vspace{5pt}
\parhead{Inverse edges.}
We enrich our \g graphs with inverse edges. Recent work has proven improved performance by adding inverse edges to ASTs \cite{hu2020}.

\vspace{5pt}
\parhead{Fine-tuning.}
After pre-training, we can fine-tune on downstream tasks. Fine-tuning involves adding downstream classifiers on top of the pre-trained node embeddings, and predicting downstream labels. A graph pooling layer \cite{wu2020} might be needed if the downstream tasks are defined on the graph/method level.

%% file: experiment.tex
\subsection{Dataset}
\begin{table}[t]
\begin{minipage}[t]{0.55\textwidth}
    \centering
    \caption{Dataset statistics. The \sigmag graph doubled the number of edges as the \mug graph, providing extra information for code graphs.}
    \label{data-statistics}
    \scalebox{0.95}{
    \begin{tabular}{ccccc}
        \toprule
        \textbf{Dataset} & \textbf{Repository} & \textbf{Method} & \textbf{Node} & \textbf{Edge}  \\
        \hline
        Java ($\sigma_0$) & 28K & 29M & 621M & 1,887M \\ 
        Java ($\sigma_1$) & 28K & 29M & 529M & 3,782M \\ 
        Python &  14K & 450K & 57M & 156M \\
        \bottomrule
    \end{tabular}
    }
\end{minipage}
\hfill
\begin{minipage}[t]{0.38\textwidth}
    \centering
        \caption{Result for method name prediction on Java dataset. Higher value indicates better performance.}
    \label{exp-mnp-java}
    \scalebox{0.95}{
    \begin{tabular}{cccc}
        \toprule
        & \textbf{F1} & \textbf{Precision} & \textbf{Recall} \\
        \hline
        \mug & 21.7 & 26.1 & 19.9 \\
        \sigmag & \textbf{23.6} & \textbf{27.5} & \textbf{22.0} \\
        \bottomrule
    \end{tabular}}
\end{minipage}
\end{table}
We tested on two real-world datasets, spanning over two programming languages Java and Python. Summary of the datasets is listed in Table \ref{data-statistics}.

\parhead{Java} 
The Java dataset is extracted from 27,581 GitHub packages. In total, these packages contain 29,024,142 Java methods. We convert each Java method into a \mug graph and a \sigmag graph. The data split is on package-level, with training (80\%), validation (10\%), and testing (10\%). 

\parhead{Python}
The Python dataset is collected from Stanford Open Graph Benchmark (\verb+ogbg-code+) \cite{hu2020}. The total number of Python methods is 452,741, with each method is represented as an AST. These ASTs are further augmented with next-token edges and inverse edges. The data split keeps in line with \verb+ogbg-code+.

\subsection{Experimental Setup}
We tune hyperparameters on all models based on their validation performances. For Java dataset, we consider a two-layer RGCN with 300 hidden units. For Python dataset, we follow \verb+ogbg-code+ and use a five-layer RGCN with 300 hidden units.
We use Adam \cite{kingma2014} as optimizer, with learning rate ranges from 0.01 to 0.0001. Mini-batch training is adopted to enable training on very large graphs\footnote[2]{\url{https://github.com/dmlc/dgl/tree/master/examples/pytorch/rgcn-hetero}}. We apply dropout at rate 0.2, L2 regularization with parameter 0.0001. The model is first pre-trained on a maximum of 10 epochs, then fine-tuned up to 100 epochs on downstream tasks until convergence. The model is implemented using Deep Graph Library (DGL) \cite{wang2019}. Models have access to 4 Tesla V100 GPUs, 32 CPUs, and 244GB memory.

\begin{figure*}[ht]
    \centering
    \begin{minipage}{.46\textwidth}
        \centering
    	\includegraphics[width=0.88\linewidth]{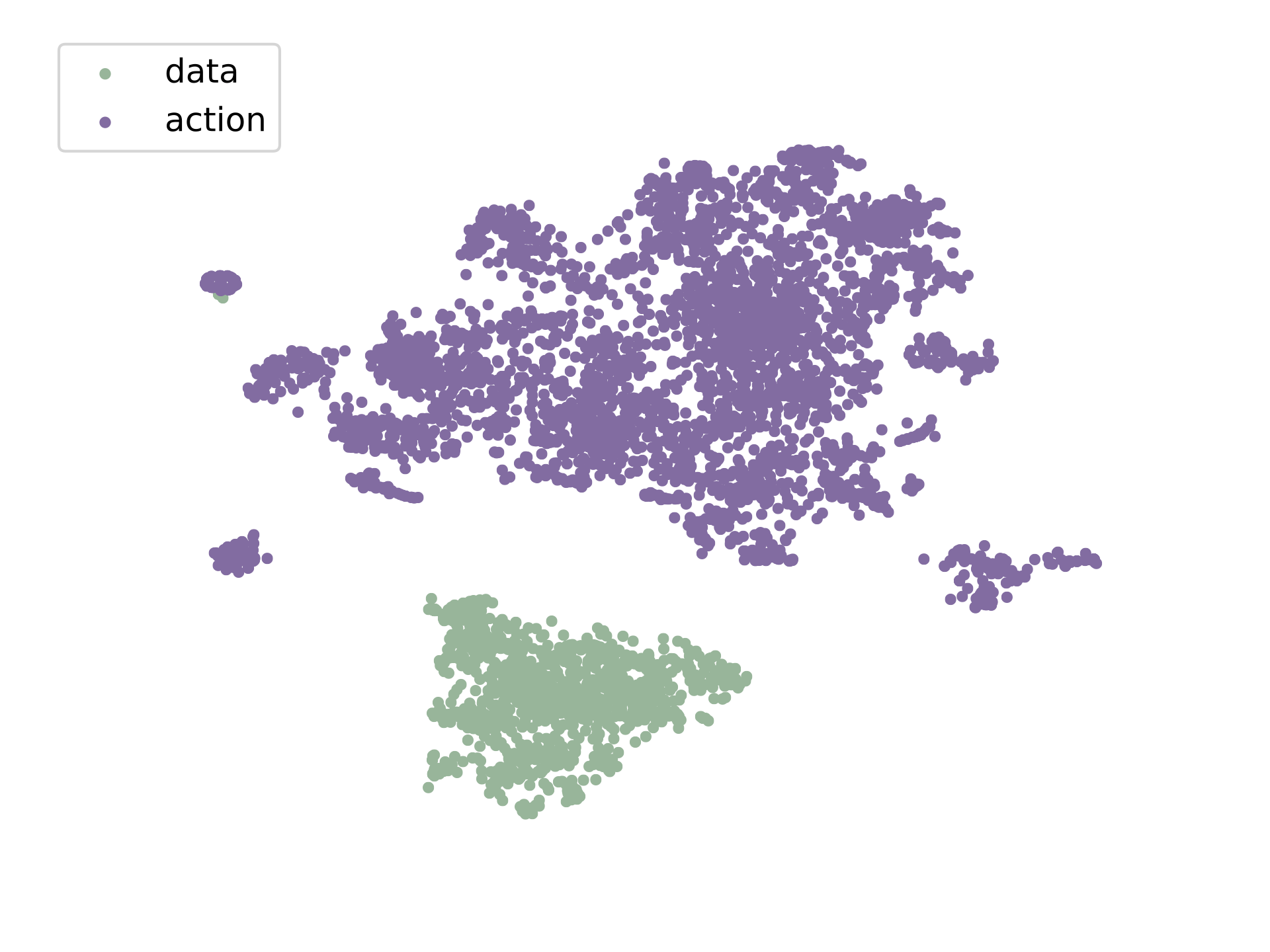}
    	\caption{Visualization of node-level embeddings with t-SNE.}
    	\label{exp-tsne-var}
    \end{minipage}%
    \hfill
    \begin{minipage}{0.5\textwidth}
        \centering
    \includegraphics[width=0.85\textwidth]{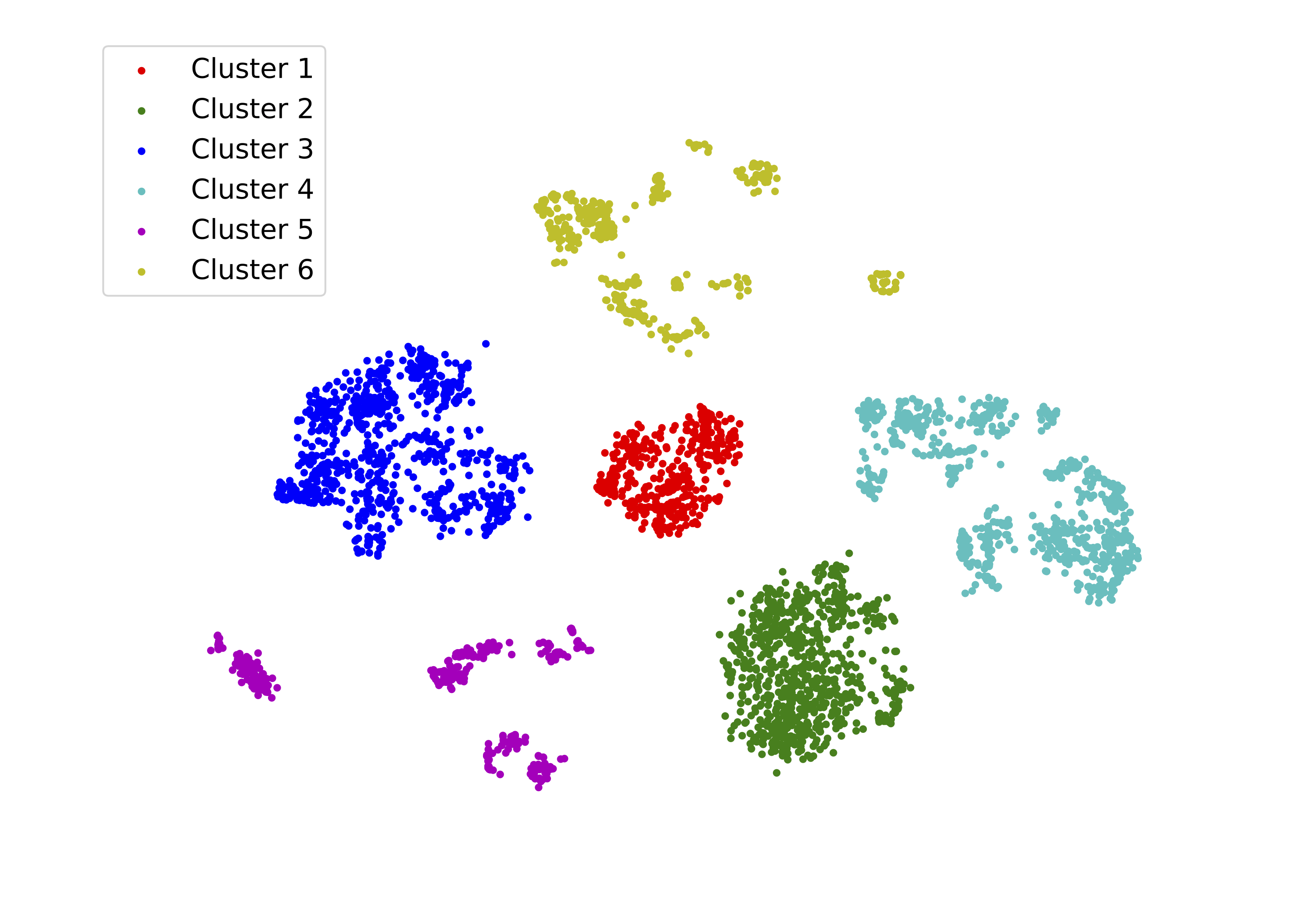}
    \caption{Visualization of method-level embeddings with t-SNE. Colored with K-means.}
\label{exp-tsne-method}
    \end{minipage}
\end{figure*}

\subsection{Analysis of Embeddings}
We begin by analyzing code embeddings via t-SNE \cite{maaten2008} visualization. We study two levels of embeddings: node-level embeddings and method-level embeddings. Here, a method-level embedding summarizes a method, computed by averaging node embeddings in its \g graph. For better visualization, we show results on 10 random Java packages (involving 4,107 Java methods) using \sigmag graphs.

In Fig. \ref{exp-tsne-var}, we see that data nodes and action nodes are forming separate clusters, indicating our code embeddings preserve important node type information.
Fig.~\ref{exp-tsne-method} suggests method embeddings are forming discriminative clusters. By manually annotating each cluster, we discovered that cluster 4 contains 91\% (out of all) \verb+set+ functions, cluster 3 contains 78\% \verb+find+ functions, and cluster 1 contains 69\% functions which end with \verb+Exception+. This result suggests that our model has the potential to distinguish methods in terms of method functionalities.

\subsection{Method Name Prediction}
We use pre-trained UniCoRN model to initialize code embeddings. Then following \cite{hu2020,alon2018}, we predict method names as downstream tasks. The method name is treated as a sequence of subtokens (e.g. \verb+getItemId+ $\rightarrow$ [\verb+get+, \verb+item+, \verb+id+]). As in \cite{hu2020}, we use independent linear classifiers to predict each subtoken. The task is defined on the method-level: predict one name for one method (code graph). We use attention pooling \cite{li2015} to generate a single embedding per method. We follow \cite{hu2020,alon2018} to report F1, precision, recall for evaluation. Below we show results on Java and Python separately, as they are used for different testing purposes.

\parhead{Java.} 
We evaluate the performance gain achieved by switching from the \mug graph to the \sigmag graph. We truncate subtoken sequences to a maximal length of 5
\begin{wrapfigure}{r}{0.32\textwidth}
\vspace{-8mm}
  \begin{center}
    \includegraphics[scale=0.5]{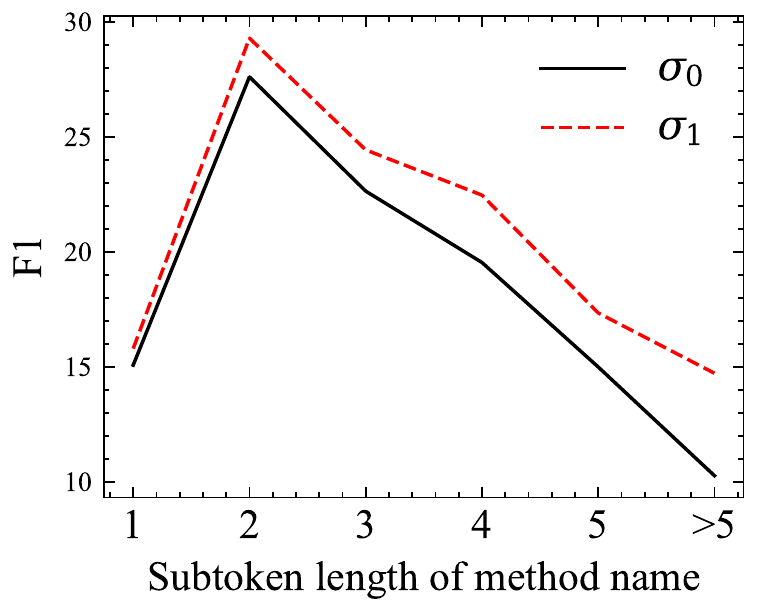}
  \end{center}
  \vspace*{-5mm}
  \caption{F1 at name length.}
  \vspace*{-5mm}
  \label{exp-mnp-java-curve}
\end{wrapfigure}
to cover 95\% of the method names. Vocabulary size is set to 1,000, covering 95\% of tokens. Tokens not in the vocabulary are replaced by a special \verb+unknown+ token. Similar techniques have been adopted in \cite{hu2020}. We experiment on approximately 774,000 methods from randomly selected 1,000 packages.

The result is summarized in Table \ref{exp-mnp-java}. We see that the \sigmag graph outperforms the \mug graph, indicating that the extra information provided by the \sigmag graph is beneficial for abstracting code snippets. Fig. \ref{exp-mnp-java-curve} further supports this observation. The \sigmag graph consistently outperforms the \mug graph w.r.t. the F1 score for different method name lengths. Note that the F1 score at method names of length 1 is low. We suspect that some names at this length are not semantically meaningful, such as \verb+a+ or \verb+xyz+. Thus, these method names are hard to predict correctly.

\begin{table}[t!]
\begin{minipage}[t]{0.535\textwidth}
    \centering
    \caption{Method name prediction for Python. Pooling: average$^\dag$, virtual node$^\S$, and attention$^\ddag$. GCN$^{\dag, \S}$ and GIN$^{\dag, \S}$: Reported in \cite{hu2020}.}
    \label{exp-mnp-python}
    \scalebox{0.86}{
    \begin{tabular}{cccc}
        \toprule
         & \textbf{F1} & \textbf{Precision} & \textbf{Recall} \\
        \hline
        GCN-NextTokenOnly$^\dag$ & 29.77 & 31.09 & 29.18 \\
        GIN-NextTokenOnly$^\dag$ & 29.00 & 30.98 & 28.13 \\
        GCN$^\dag$ & 31.63 & - & - \\
        GIN$^\dag$ & 31.63 & - & - \\
        UniCoRN w/o pretrain$^\dag$ & 32.81 & 35.25 & 31.71 \\
        UniCoRN$^\dag$ & 33.28 & 35.28 & 32.36 \\
        \hline
        GCN$^\S$ & 32.63 & - & - \\
        GIN$^\S$ & 32.04 & - & - \\
        UniCoRN$^\S$ & 33.80 & 35.81 & 32.89 \\
        \hline
        GCN$^\ddag$ & 32.80 & 34.72 & 31.88 \\
        GIN$^\ddag$ & 32.60 & 34.42 & 31.77 \\
        UniCoRN$^\ddag$ & \textbf{33.94} & \textbf{36.02} & \textbf{32.99} \\
        \bottomrule
    \end{tabular}}
\end{minipage}
\hfill
\begin{minipage}[t]{0.43\textwidth}
    \caption{MRR and Hit@K(\%) results for link prediction.  Higher values are better. Superscripts $^\calD$ and $^\calM$ denote DistMult and MLP link predictors. Hit@K for random is computed as $K/(1+200)$, where 200 is the number of negative edges per testing edge.}
    \label{exp-lp-java}
    \centering
    \scalebox{0.86}{
    \begin{tabular}{ccccc}
        \toprule
         & \textbf{MRR} & \textbf{Hit@1} & \textbf{Hit@3} & \textbf{Hit@10} \\
        \hline
        Random & - & 0.5 & 1.5 & 5.0 \\
        \hline
        FastText$^\calD$ & 0.01 & 0.4 & 1.0 & 2.3\\
        $\sigma_0^\calD$ & 0.26 & 15.4 & 28.5 & 41.3 \\
        $\sigma_1^\calD$ & \textbf{0.32} & \textbf{18.1} & \textbf{38.4} & \textbf{61.4} \\
        \hline
        FastText$^\calM$ & 0.05 & 1.9 & 4.0 & 8.0 \\
        $\sigma_0^\calM$ & 0.51 & 46.1 & 49.8 & 58.4 \\
        $\sigma_1^\calM$ & \textbf{0.53} & \textbf{46.2} & \textbf{55.0} & \textbf{65.2} \\
        \bottomrule
    \end{tabular}
    }
\end{minipage}
\end{table}

\parhead{Python.} 
We compare the performances of UniCoRN with various baselines on Python. Our experiment setup closely follows \verb+ogbg-code+ \cite{hu2020}. For the baseline, \verb+ogbg-code+ considers GCN and GIN. Additionally, we introduce two baselines that run GCN and GIN with next-token edges only. We expect these two new baselines to mimic the performance of sequence-based models. In this task, we test three pooling methods: average, virtual node \cite{hu2020}, and attention \cite{li2015}.

Results are given in Table \ref{exp-mnp-python}. We list three observations. First, UniCoRN with attention pooling (UniCoRN$^\ddag$) performs the best, endorsing UniCoRN's superior modeling capacity. Second, UniCoRN with pre-training shows performance gain over UniCoRN without pre-training, verifying the usefulness of our pre-training strategies. Third, GCN(GIN) improves GCN(GIN)-NextTokenOnly, confirming the importance of using structural information.

\begin{figure}[t]
\begin{minipage}[t]{0.4\textwidth}
    \centering
    \scriptsize
    \begin{tabular}{cc}
        \toprule
        \textbf{Ground truth} & \textbf{Prediction} \\
        \hline
         \verb+get_config+ & \verb+get_config+ \\
         \verb+create_collection+ & \verb+create_collection+\\
         \verb+get_aws_credentials+ & \verb+get_ec2+\\
         \verb+wait_for_task_ended+ & \verb+wait_job+\\
        \verb+add_role+ & \verb+create_permission+\\
         \verb+load_bytes+ & \verb+upload_file+\\
        \bottomrule
    \end{tabular}
    \caption{Examples of method name prediction on Python in different degree of consensus. Each pair of results is demonstrated as ground truth name and predicted name.}
    \label{exp:mnp-pre}
\end{minipage}
\hfill
\begin{minipage}[t]{0.54\textwidth}
    \centering
    \vspace*{-12mm}
        \begin{minted}[fontsize=\scriptsize]{python}
def wait_for_task_ended(self):
    try:
        waiter = self.client.\
            get_waiter('job_execution_complete')
        # timeout is managed by airflow
        waiter.config.max_attempts = sys.maxsize  
        waiter.wait(jobs=[self.jobId])
    except ValueError:
        # If waiter not available use expo
        retry ...
    \end{minted}
    Prediction: \verb+wait_job+.
    \cprotect\caption{Reasonable prediction based on the code context is observed, though it is inexact match.}
\label{exp:mnp-code-pre}
\end{minipage}
\end{figure}

Example pairs of ground truth and prediction are shown in Fig. \ref{exp:mnp-pre}. The examples of prediction encompass exact matches, such as \verb+get_config+ pair, context matches, such as \verb+get_aws_credentials+ pair, and mismatches, such as \verb+load_bytes+ pair. We showcase a prediction example in Fig. \ref{exp:mnp-code-pre}.

\subsection{Link Prediction}
In this task, we examine how UniCoRN recovers links in code graphs. We follow the same experimental setup as in \cite{ioannidis2020}. We feed two node embeddings of a link to a predictor, a DistMult \cite{yang2014} or a two-layer MLP, to predict the existence of the link. Here, node embeddings are obtained by applying UniCoRN to $\sigma_0$/\sigmag graphs, or simply obtained from FastText embeddings. Note that the FastText is the initial embedding of UniCoRN. We freeze UniCoRN after pre-training. When fine-tuning link predictors, we ensure \mug and \sigmag have the same set of training edges. During inference, we sample 200 negative edges per testing edge (positive) and evaluate the rank of the testing edge. Evaluations are based on Mean Reciprocal Rank (MRR) and Hit@K (K=1, 3, 10). 

\begin{wrapfigure}{r}{0.35\textwidth}
\vspace{-10mm}
  \begin{center}
    \includegraphics[scale=0.5]{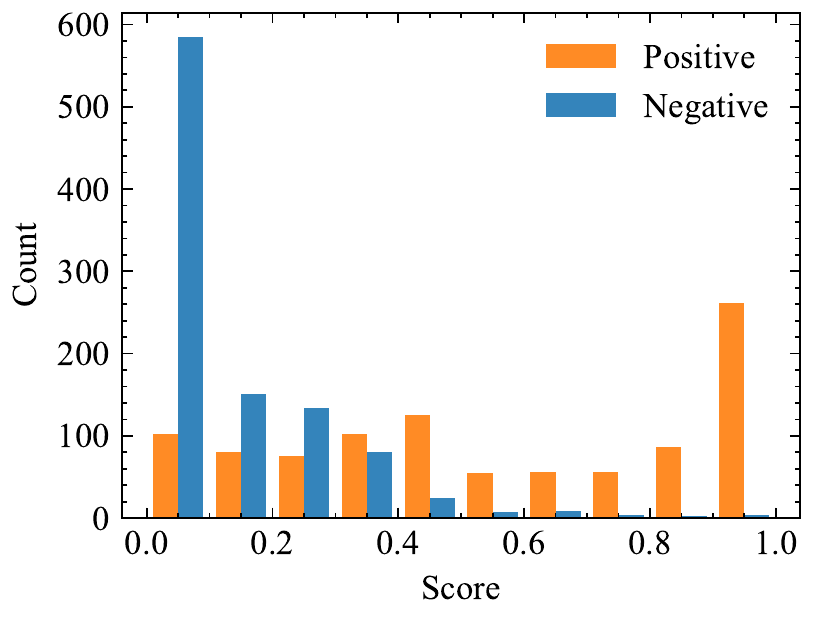}
  \end{center}
  \vspace*{-5mm}
  \caption{Scores for positive and negative edges.}
  \label{exp-lp-histogram-java}
  \vspace*{-7mm}
\end{wrapfigure}
Table \ref{exp-lp-java} shows the results. UniCoRN outperforms FastText. The results suggest that node embeddings from UniCoRN capture neighborhood correlations. We see \sigmag graph again improves \mug graph. Fig. \ref{exp-lp-histogram-java} demonstrates the histogram of scores for 1,000 positive and 1,000 negative edges. The score, which ranges from 0 to 1, indicates the plausibility of the link existence. Positive edges (0.58$\pm$0.33) score higher than negative edges (0.13$\pm$0.15), with $p$-value less than 0.00001 using \textit{t}-test, suggesting that UniCoRN is capable to distinguish positive and negative links in code graphs.

%% file: conclusion.tex
This paper presents a new model, UniCoRN, to provide a universal representation for code. Building blocks of UniCoRN include a novel \g graph to represent code as graphs, and four effective signals to pre-train GNNs. Our pre-training framework enables fine-tuning on various downstream tasks. Empirically, we show UniCoRN's superior ability to offer high-quality code representations.

There are several possibilities for future works. First, we are looking to enhance UniCoRN with additional code-specific signals. Second, the explainability of the learned code representation deserves further study. The explainability can in turn motivate additional signals to embed desired code properties. Third, more downstream applications are left to be explored, such as bug detection and duplicate code detection upon the availability of labeled data.